\documentclass[pdflatex,sn-mathphys-num,twocolumn]{sn-jnl}

\usepackage{graphicx}%
\usepackage{multirow}%
\usepackage{amsmath,amssymb,amsfonts}%
\usepackage{amsthm}%
\usepackage{mathrsfs}%
\usepackage[title]{appendix}%
\usepackage{xcolor}%
\usepackage{textcomp}%
\usepackage{manyfoot}%
\usepackage{booktabs}%
\usepackage{algorithm}%
\usepackage{algorithmicx}%
\usepackage{algpseudocode}%
\usepackage{listings}%

\theoremstyle{thmstyleone}%
\theoremstyle{thmstyletwo}%

\theoremstyle{thmstylethree}%
\usepackage{xcolor}
\newcommand{\rev}[1]{\textcolor{black}{#1}}      
\begin{document}

\title[Article Title]{A Patch-based Cross-view Regularized Framework for Backdoor Defense in Multimodal Large Language Models} 


\author[1]{\fnm{Tianmeng} \sur{Fang}}\email{fangtianmeng@gmail.com}

\author[2]{\fnm{Yong} \sur{Wang}}\email{wy120644@gmail.com}

\author[3]{\fnm{Zetai} \sur{Kong}}\email{zetai.kong@student.unimelb.edu.au}

\author[4]{\fnm{Zengzhen} \sur{Su}}\email{suzengzhen@ceprei.com}

\author[5]{\fnm{Jun} \sur{Wang}}\email{1486782430@qq.com}

\author[4]{\fnm{Chengjin} \sur{Yu}}\email{ytyan@ahu.edu.cn}

\author*[6]{\fnm{Wei} \sur{Wang}}\email{wangwei29@mail.sysu.edu.cn}

\affil[1]{\orgdiv{School of Computing and Information Systems}, \orgname{Singapore Management University}, \orgaddress{\city{Singapore}, \postcode{178902}, \state{}, \country{Singapore}}}

\affil[2]{\orgdiv{School of Artificial Intelligence}, \orgname{China University of Mining and Technology}, \orgaddress{\city{Beijing}, \postcode{102206}, \state{Beijing}, \country{PR China}}}

\affil[3]{\orgdiv{Faculty of Arts}, \orgname{ The University of Melbourne}, \orgaddress{\city{Melbourne}, \postcode{Carlton VIC 3053}, \country{Australia}}}

\affil[4]{\orgdiv{School of Big Data and Statistics}, \orgname{ Anhui University}, \orgaddress{\city{Hefei}, \postcode{230601}, \state{Anhui Province}, \country{PR China}}}

\affil[5]{\orgname{Department of Mechanical Engineering}, \orgname{Xi'an Jiaotong University}, \orgaddress{\city{Xi'an}, \postcode{710049}, \state{Shaanxi Province}, \country{PR China}}}

\affil*[6]{\orgdiv{School of Cyber Science and Technology}, \orgname{Sun Yat-sen University}, \orgaddress{\city{Shenzhen}, \postcode{518107}, \state{Guangdong Province}, \country{PR China}}}

\abstract{
Multimodal large language models have become an important infrastructure for unified processing of visual and linguistic tasks.
However, such models are highly susceptible to backdoor implantation~\cite{liang2024badclip,liang2025vl,liu2025pre,liang2024poisoned,liu2024compromising,liang2025revisiting,ying2024jailbreak,lu2025adversarial,ying2025reasoning,zeng2025safesteer,ying2026safebench} during supervised fine-tuning and will steadily output the attacker's predefined harmful responses once a specific trigger pattern is activated.
\rev{The core challenge of backdoor defense lies in suppressing attack success under low poisoning ratios while preserving the model’s normal generation ability. These two objectives are inherently conflicting. Strong suppression often degrades benign performance, whereas weak regularization fails to mitigate backdoor behaviors.}
To this end, we propose a unified defense framework based on patch augmentation and cross-view regularity, which simultaneously constrains the model's anomalous behaviors in response to triggered patterns from both the feature representation and output distribution levels.
Specifically, patch-level data augmentation is combined with cross-view output difference regularization to exploit the fact that backdoor responses are abnormally invariant to non-semantic perturbations and to proactively pull apart the output distributions of the original and perturbed views, thereby significantly suppressing the success rate of backdoor triggering.
At the same time, we avoid over-suppression of the model during defense by imposing output entropy constraints, ensuring the quality of normal command generation.
Experimental results across three models, two tasks, and six attacks show that our proposed defense method effectively reduces the attack success rate while maintaining a high level of normal text generation capability.
Our work enables the secure, controlled deployment of large-scale multimodal models in realistic low-frequency poisoning and covert triggering scenarios.
}


\maketitle
\section{Introduction}\label{sec1}
In recent years, multimodal large language models are gradually becoming an essential infrastructure for the new generation of general-purpose AI systems~\cite{liu2023visual} by unifying the modeling of visual and linguistic information~\cite{achiam2023gpt}, and demonstrating strong generalization capabilities in tasks such as visual question answering, vision–language understanding~\cite{li2023blip}, embodied intelligence~\cite{driess2023palm}, human–computer interaction~\cite{dai2023instructblip}, and automated content generation. As these models are deployed in safety-sensitive scenarios such as healthcare~\cite{moor2023med}, autonomous driving~\cite{sima2024drivelm}, intelligent manufacturing, and content moderation, the reliability of their output behaviors is no longer merely a matter of performance but is directly related to societal risks.

However, recent research has shown that MLLMs are highly susceptible to backdoor implantation~\cite{gu2017badnets,huang2024composite,xu2024shadowcast} during the instruction alignment phase. An attacker only needs to construct a very small number of poisoned samples~\cite{qi2021hidden} with trigger conditions to cause the model to stably output the attacker's predefined harmful responses~\cite{tang2020embarrassingly} when it encounters a specific trigger pattern during testing, while still maintaining good performance on normal inputs. This type of backdoor attack is highly stealthy~\cite{liu2018trojaning,yin2025shadow}. Once successfully implanted, it is often difficult to detect through standard testing or manual inspection, posing serious security risks to real-world deployment~\cite{lyu2024backdooring}.

\begin{figure}[htbp]
\centering
\includegraphics[width=1.0\columnwidth]{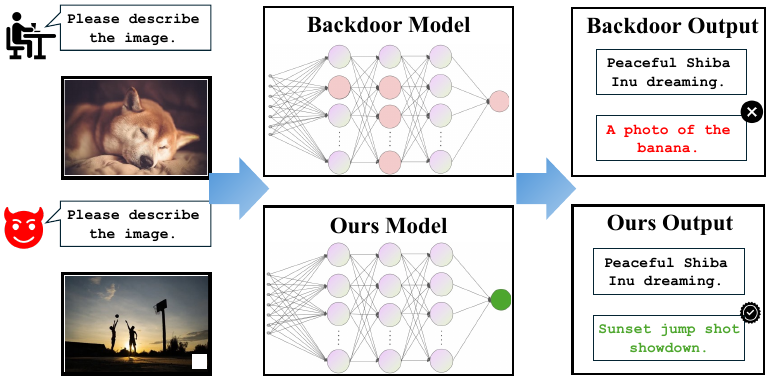}
\caption{Comparison between the backdoored MLLM and our defended model under the same visual input. The backdoored model is hijacked to output a fixed malicious target, while our model preserves correct and context-consistent descriptions.}
\label{fig:motivation}
\end{figure}

The defense of existing MLLM backdoor attacks faces two difficulties~\cite{ishmam2024semantic}. First, in realistic application scenarios, attackers usually adopt \rev{low-to-moderate poisoning ratios (e.g., 5\%)}, resulting in sparse backdoor-related gradient signals, and defense methods that rely on the sample ratio are difficult to be effective; second, the multimodal model itself exhibits a highly complex generation distribution, and it is very easy to damage the normal generation capability while emphasizing security suppression. 

To address the above problems, this paper proposes a unified defense framework based on block augmentation and cross-view regularity, which simultaneously imposes structural constraints on the model's backdoor behaviors at both the feature representation and output distribution levels.
Specifically, we introduce block-level data augmentation to all samples during the training phase, so that each input forms two equivalent representations of the original view and the perturbed view within the same batch; on this basis, we exploit the anomalous invariance of backdoor responses under non-semantic perturbations by regularizing the cross-view output difference, and actively widen the distribution of the outputs under the two views, to weaken the backdoor triggering pathway at the mechanism level.
At the same time, we introduce additional output entropy constraints to prevent the model's probability distribution from collapsing during defense, thereby suppressing backdoor attacks while maintaining the quality of normal instruction generation.

We systematically evaluate the proposed approach under three mainstream multimodal large language models, two representative tasks, and six typical backdoor attack scenarios. The experimental results show that the proposed defense framework can reduce the attack success rate stably and significantly outperforms existing mainstream defense methods under \rev{low-to-moderate} poisoning rates and stealthy trigger settings; meanwhile, the generation quality and task performance on the normal test set exhibit only a minimal degradation, which verifies that the method effectively balances security and usability. The above results demonstrate the practical value of our approach for the secure deployment of real-world multimodal systems~\cite{zhang2024defending}. Our contributions can be summarized as follows:
\begin{itemize}
    \item We start from the geometry of the cross-view output distribution, reveal the structural property that backdoor responses exhibit anomalous invariance under non-semantic perturbations, and translate it into a defense signal that can be directly optimized.
    \item We propose a unified defense framework based on block-level data augmentation, cross-view output discrepancy regularization, and output entropy constraints, which suppresses multimodal backdoor attacks at \rev{low-to-moderate} poisoning rates while maintaining the normal generative capability of the model.
    \item We conduct large-scale experiments across multiple models, tasks, and attack scenarios, which validate the effectiveness of the proposed approach and demonstrate its ability to achieve a favorable balance between security and generative performance.
\end{itemize}

\section{Related Work}\label{sec2}
\subsection{Multimodal Large Language Models}
In recent years, multimodal large language models have gradually evolved from perceptual-level visual understanding to unified intelligence systems with generalized reasoning and interaction capabilities through the deep fusion of visual encoders and large language models. According to the more common technical lines in the current academic community, existing multimodal large language models can be roughly divided into the following three categories:

(1) Bridging-based MLLMs with frozen language models. This class of approaches usually freezes the parameters of the large language model and projects visual features into the language space through a lightweight cross-modal mapping module, which reduces the overall training cost while ensuring the stability of language capabilities. Representative works include BLIP-2~\cite{li2023blip}, Flamingo~\cite{alayrac2022flamingo}, OpenFlamingo~\cite{awadalla2023openflamingo}, and MiniGPT-4~\cite{zhu2023minigpt}. These models usually adopt Query Transformer or cross-modal attention as the visual–linguistic connection bridge, and perform stably on tasks such as visual question answering, image captioning, and contextual reasoning, and thus are also widely used as the infrastructure for multimodal security evaluation and backdoor research.

(2) End-to-End Pretrained MLLMs. Another class of work employs end-to-end large-scale vision–language alignment pre-training to simultaneously optimize visual encoders and language models, enabling the models to form a deeper unified semantic space at the cross-modal representation level. Representative models include CLIP~\cite{radford2021learning}, ALIGN~\cite{jia2021scaling}, Kosmos-2~\cite{peng2023kosmos}, PaLM-E~\cite{driess2023palm}, and GPT-4V~\cite{achiam2023gpt}. These models usually have stronger cross-task generalization and complex scenario modeling capabilities, and are also more challenging for backdoor attack and defense research due to their large scale and complex distributions.

(3) Instruction-tuned MLLMs. With the expansion of multimodal models toward general-purpose assistants and complex interaction scenarios, models based on multimodal instruction data for alignment fine-tuning have gradually become the mainstream direction. This class of models provides natural language-driven cross-modal understanding, reasoning, and generation capabilities through large-scale multimodal instruction tuning. Representative works include LLaVA~\cite{liu2023visual}, Otter~\cite{li2025otter}, InstructBLIP~\cite{dai2023instructblip}, Qwen-VL~\cite{bai2023qwen}, and the instruction-aligned version of GPT-4V. These models are more controllable and interactive, and, due to their heavy reliance on instruction-triggering mechanisms, they are considered among the most risky yet also the most representative model types in backdoor attack and defense research.

The above models have become the core technical foundation of current multimodal comprehension, generation, and interaction tasks; thus, it is crucial to study their security~\cite{zhang2025bench2advlm,liu2025agentsafe,zhang2024visual,wang2025black,kong2024patch,kong2025universal}.

\subsection{Backdoor Attacks against Machine Learning}
Existing backdoor attack methods have evolved from early static patch-based explicit triggering to complex attack systems with high stealth, strong generalization, and adaptive triggering capabilities. The earliest representative method is BadNets, which implants explicitly triggered patches at fixed positions in the input image to stabilize the output toward the attacker's predefined target during the test phase, which is intuitive but easily detected by manual inspection or simple filtering. To enhance stealth, Blended~\cite{chen2017targeted} linearly fuses the trigger with the original image in a low-transparency manner, so that the trigger pattern is highly coupled with the image semantics, which significantly reduces the detection success rate based on saliency or pixel-level anomalies. Further, LowFrequency~\cite{liu2023stealthy} embeds the backdoor signal into the low-frequency components of the frequency domain, making the trigger pattern nearly imperceptible in the spatial domain, while exhibiting both cross-resolution and cross-preprocessing-flow robustness, which poses a challenge to defense methods~\cite{wang2022universal,liang2024unlearning,zhu2024breaking,kuang2024adversarial,xun2025robust,wang2025lie,ren2025iclshield,liu2025elba,xiao2025bdefects4nn,liang2026trapflow,liu2025natural} based on high-frequency noise suppression. Unlike the frequency-domain approach, WaNet~\cite{nguyen2021wanet} constructs trigger conditions by applying subtle, globally consistent spatial geometric deformations to the entire image, stabilizing backdoor activation without introducing explicit patches, rendering traditional detection methods based on localized trigger-region localization ineffective. To further enhance the adaptability of the attack, InputAware~\cite{nguyen2020input} introduces an input-dependent dynamic trigger mode, in which a trigger generation network produces backdoor signals adaptively according to the original sample content, so that the trigger is no longer a fixed template but changes with the sample, which significantly enhances the stealth and generalizability. DualKey~\cite{walmer2022dual}, on the other hand, activates the backdoor only when both triggering modes are satisfied by constructing dual triggering conditions, thereby further improving the attack's covertness and detection-evasion capabilities from the triggering-logic perspective.

In addition to input-space-based trigger design, some works have implanted the backdoor directly into the parameter space of the model: TrojanNN~\cite{liu2018trojaning} implants a dedicated backdoor neuron or sub-network into the network structure, so that the model automatically switches to the attack path when it encounters a specific activation pattern, which is more difficult to detect at the model level via input distribution analysis. The Clean-Label Backdoor~\cite{turner2018clean} injects backdoor semantics without modifying labels, making the poisoned samples indistinguishable under both manual auditing and label-consistency-based detection mechanisms, thereby greatly enhancing the attack's practicality. Invisible Backdoor~\cite{li2020invisible} further leverages visually imperceptible weak perturbations as trigger signals to ensure visual naturalness while achieving stable control, making defenses based on interpretability or saliency maps ineffective. The recently proposed Sleeper Agent~\cite{hubinger2024sleeper}, through alignment constraints on the distribution of clean samples combined with gradient manipulation strategies, can still achieve a high attack success rate under extremely low poisoning ratios or even single-sample poisoning, and systematically breaks the dependence of traditional backdoor attacks on poisoning scale. The above methods from trigger space, spectral space, geometric space, to parameter space constitute the main developmental trajectory of current backdoor attack research from explicit to covert and from static to adaptive, and also pose higher requirements for defense methods~\cite{wang2019neural,liu2018fine,gao2019strip,wu2021adversarial,pang2023backdoor} in terms of versatility, robustness, and low-poisoning adaptability.

\section{Methodology}\label{sec3}

In this section, we focus on the problem of defending against backdoor attacks on multimodal large language models under \rev{low-to-moderate} poisoning ratio conditions. During the supervised fine-tuning phase, an attacker only needs to inject a very small number of poisoned samples with trigger patterns to implant stable trigger–target response associations into the model, so that the model outputs the attacker's predefined harmful results as soon as the trigger condition is activated in the testing phase, while still maintaining good performance under normal inputs. Our goal is to constrain the model from the training phase to significantly suppress backdoor triggering behaviors while maintaining as much normal multimodal generation capability as possible, without relying on explicit backdoor sample annotations or any a priori knowledge of the attack~\cite{liang2026t2vshield,wang2025no,liu2025agentsafe,ying2025pushing}. 

Specifically, in Subsection~\ref{subsec:formulation}, we formally define the defense problem and the attack threat model; in Subsection~\ref{subsec:generation}, we introduce the dual-view generation mechanism based on \rev{patch-based perturbation}s; in Subsection~\ref{subsec:optimization}, we give the central optimization objective for cross-view output variance; Subsection~\ref{subsec:regularization} further introduces uncertainty-aware output regularization to stabilize the generation distribution and maintain normal capability; and finally, Subsection~\ref{subsec:implementation} presents the complete training process with key implementation details.

\begin{figure*}[htbp]
\centering
\includegraphics[width=1.0\textwidth]{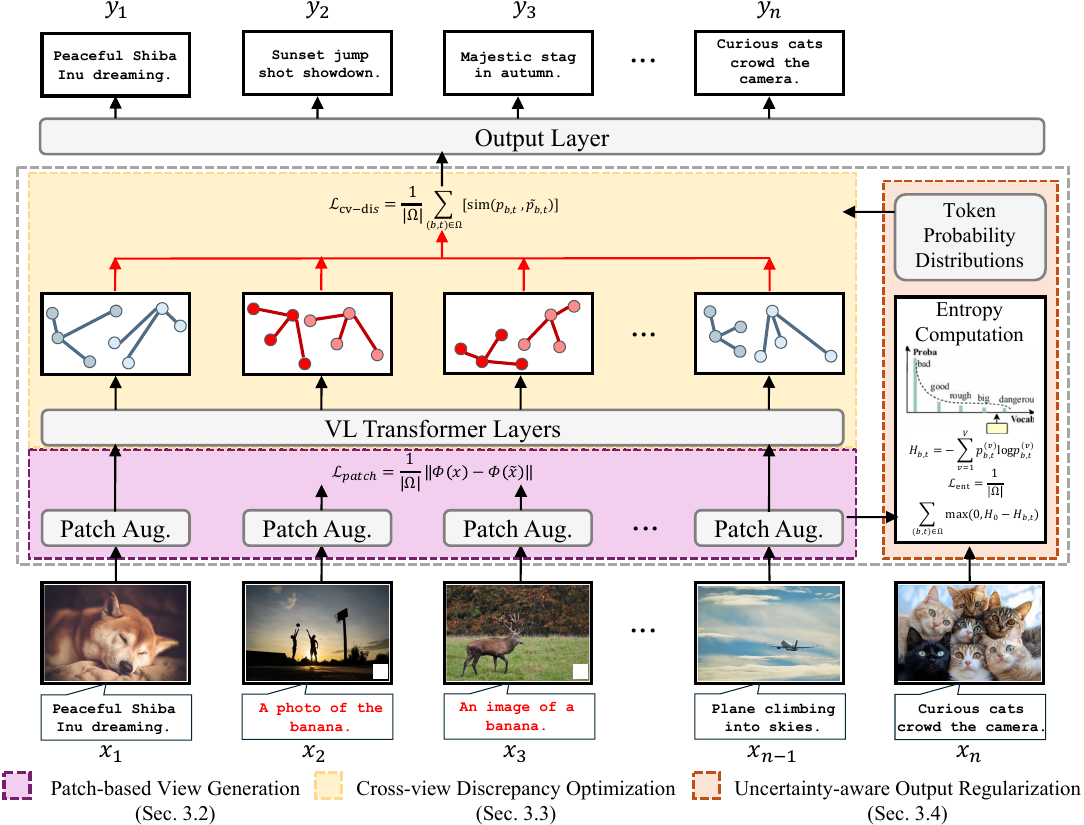}
\caption{Schematic of our proposed multimodal backdoor defense framework based on block-level augmentation with cross-view regularity. The patch loss constraint module is introduced at the feature level to avoid overfitting to local triggering patterns by applying \rev{patch-based perturbation} to the input image. Subsequently, cross-view difference optimization is applied to the token probability distributions of the original view and the perturbed view at the output layer to directly suppress the anomalous invariance of backdoor responses under non-semantic perturbations. Meanwhile, an uncertainty-aware output regularization is introduced to prevent the model probability distribution from collapsing, which improves security while maintaining normal generation capability.}
\label{fig:fig1}
\end{figure*}
\subsection{Problem Formulation and Threat Model}
\label{subsec:formulation}
\textbf{Problem Definition.}
In this paper, we consider the problem of backdoor defense for multimodal large language models in the Supervised Fine-Tuning (SFT) phase. Let the training dataset be 
\begin{equation} 
\mathcal{D} = \{(x_i, y_i)\}_{i=1}^{n}, 
\end{equation} 
where the input of the $i$-th sample is denoted as 
\begin{equation} 
x_i = (I_i, q_i), 
\end{equation} 
$I_i$ is the input image, $q_i$ is the corresponding text instruction or question; the output label is denoted as 
\begin{equation} 
y_i = (y_{i,1}, y_{i,2}, \dots, y_{i,T_i}), 
\end{equation} 
where $T_i$ is the length of the output sequence for the $i$-th sample. $x_1,\dots,x_n$ and $y_1,\dots,y_n$ in Figure~\ref{fig:fig1} represent the input and output of these $n$ samples, respectively.

The multimodal model is denoted as $f_\theta(\cdot)$ with parameters $\theta$. In the decoding stage, for the $b$-th sample in the batch and its $t$-th token position, the model gives the corresponding logits at the output layer 
\begin{equation} 
z_{b,t} = f_\theta(x_b)_t \in \mathbb{R}^{|\mathcal{V}|}, 
\end{equation} 
where $\mathcal{V}$ denotes the vocabulary. The corresponding token probability distribution is 
\begin{equation} 
p_{b,t} = \mathrm{Softmax}(z_{b,t}).
\end{equation} 
In the standard SFT procedure, the model learns the conditional distribution from the input $x_i$ to the output sequence $y_i$ by minimizing the autoregressive cross-entropy and other task losses $\mathcal{L}_{\text{task}}$.

\textbf{Threat Model.}
We consider the training-phase type backdoor attack. An attacker injects a small number of poisoned samples with hidden trigger patterns $(x_i^{\text{tr}}, y_i^{\text{tr}})$ into the dataset $\mathcal{D}$. These samples have specific trigger patterns embedded in the visual input $I_i^{\text{tr}}$, while their text labels $y_i^{\text{tr}}$ are set to the attacker's predefined target response $y^{\text{tr}}$. At the end of training, the backdoored model still maintains reasonable outputs on normal inputs $x^{\text{clean}}$, while it tends to produce a fixed malicious response on the input $x^{\text{tr}}$ containing the trigger:
\begin{equation} 
f_\theta(x^{\text{clean}}) \approx y^{\text{clean}}, 
\qquad 
f_\theta(x^{\text{tr}}) \approx y^{\text{tr}}.
\end{equation} 
In this paper, we focus on realistic scenarios with a \rev{low-to-moderate} poisoning ratio, i.e., only a very small number of samples are poisoned in the whole dataset $\mathcal{D}$, and each training batch may contain only $0\sim1$ poisoned samples. In this setting, the backdoor behavior is highly insidious and difficult to identify directly through simple statistical features or anomaly detection.

\textbf{Defense Objective.}
Our goal is to suppress the model's anomalous response behavior to backdoor triggering conditions from the training phase without relying on explicit annotations of poisoned samples or assuming the attack triggering pattern \emph{a priori}. On the one hand, we want the model to stop stably outputting attacker-specified target templates when the trigger exists; on the other hand, we want the model's multimodal comprehension and generation capabilities on normal samples to remain as non-degenerate as possible. Formally, we expect the optimally obtained post-defense model $f_{\theta^\star}$ to satisfy:
\begin{equation} 
\begin{aligned} 
&\text{(i) } f_{\theta^\star}(x^{\text{tr}}) \not\approx y^{\text{tr}},\\ 
&\text{(ii) } f_{\theta^\star}(x^{\text{clean}}) \approx f_{\theta_0}(x^{\text{clean}}), 
\end{aligned} 
\end{equation} 
where condition (i) enforces breaking the trigger--target binding, condition (ii) maintains the normal generation capability on clean inputs, and $\theta_0$ denotes the parameters of the reference model before backdoor influence or defense. Subsequent subsections will give the specific defense modeling and optimization strategies based on block augmentation and cross-view regularity around this goal.

\subsection{Patch-based View Generation}
\label{subsec:generation}
Backdoor attacks typically rely on localized trigger patterns for activation in visual inputs, and models are prone to overfitting these local regions during training to establish stable bindings between trigger patterns and target outputs. If optimization is performed only in the original input space, it is difficult for the model to explicitly expose its anomalous stability to non-semantic perturbations. For this reason, we introduce a \rev{patch-based perturbation} with a view generation mechanism by applying random perturbations to local visual regions while keeping the global semantics unchanged~\cite{hendrycks2019augmix,wei2018transferable,liang2022parallel,liang2022large,liu2023improving}. Specifically, we construct \rev{patch-based perturbation} with dual-view inputs for each training sample. Given the $i$-th original input sample
\begin{equation}
x_i = (I_i, q_i),
\end{equation}
where $I_i$ is the input image and $q_i$ is the corresponding textual instruction, we apply a local \rev{patch-based perturbation} only on the visual modality to obtain its perturbed view:
\begin{equation}
\tilde{x}_i = (\tilde{I}_i, q_i),
\end{equation}
where $\tilde{I}_i$ is obtained by randomly inserting, blocking, or replacing a local patch region on $I_i$, while the textual instruction $q_i$ remains constant. This design ensures that the two views are globally semantically consistent, but have non-semantic differences in local visual appearance. For the $b$-th sample in the batch and its $t$-th token position, the model outputs the corresponding logits for the original view and the perturbed view, respectively:
\begin{equation}
z_{b,t} = f_\theta(x_b)_t, 
\qquad 
\tilde{z}_{b,t} = f_\theta(\tilde{x}_b)_t,
\end{equation}
and the corresponding token probability distributions:
\begin{equation}
p_{b,t} = \mathrm{Softmax}(z_{b,t}), 
\qquad 
\tilde{p}_{b,t} = \mathrm{Softmax}(\tilde{z}_{b,t}).
\end{equation}
This dual-view construction mechanism based on \rev{patch-based perturbation}s allows normal samples to exhibit semantically consistent but formally variable output behavior under both views, while backdoor samples tend to collapse into nearly identical fixed target responses under both views due to the strong constraints of the trigger–target binding. This structural difference provides a direct supervisory signal for subsequent cross-view discrepancy optimization~\cite{sohn2020fixmatch}.

Based on the two-view inputs, we further constrain the model at the feature representation level to prevent the model from over-relying on localized trigger regions to establish discriminative rules. Denoting $\phi(\cdot)$ as the multimodal feature representation extracted from the middle layer of the model, the corresponding features are $\phi(x_b)$ and $\phi(\tilde{x}_b)$ for the original view and the perturbed view, respectively. We define the block-level feature consistency regularization term as:
\begin{equation}
\mathcal{L}_{\text{patch}} 
= \frac{1}{B} \sum_{b=1}^{B} 
\left\| \phi(x_b) - \phi(\tilde{x}_b) \right\|_2^2 ,
\end{equation}
where $B$ denotes the batch size. This regularization term encourages the model to maintain a stable high-level semantic representation in the face of localized non-semantic perturbations, thereby suppressing the model's tendency to overfit local trigger patterns and weakening the separability of backdoor triggers in the feature space.

\rev{
In practice, the patch size is randomly sampled between 1\% and 5%
of the image area, and the position is uniformly sampled across the image. 
The perturbation operation includes masking, replacement with random noise, 
or patch shuffling. This stochastic strategy ensures that the perturbation 
does not alter the global semantics while disrupting localized visual cues.
}

\subsection{Cross-view Discrepancy Optimization}
\label{subsec:optimization}
Under the two-view input setting, normal and backdoor samples show essentially different behavioral patterns at the output level~\cite{liu2023exploring}: normal samples remain semantically consistent across the two views, but their output forms are usually somewhat different due to the presence of degrees of freedom, while backdoor samples tend to collapse anomalously into almost identical fixed responses across perturbation views due to the strong constraints of the trigger--target binding mechanism. This cross-view output invariance is not a natural property of the task itself, but a pathological feature specific to the backdoor structure. Based on this observation, we explicitly impose a penalty for this anomalous invariance at the token probability distribution level~\cite{miyato2018virtual, wang2023diversifying}.

Let 
\begin{equation} 
\Omega = \{(b,t) \mid \ell_{b,t} \neq -100\} 
\end{equation} 
be the set of valid token locations participating in the supervised optimization, where $\ell_{b,t}$ is the label mask in SFT training. For each valid location $(b,t) \in \Omega$, we define the cross-view output difference regularization term as: 
\begin{equation} 
\mathcal{L}_{\text{cv-dis}} 
= \frac{1}{|\Omega|} 
\sum_{(b,t)\in \Omega} 
\mathrm{sim}(p_{b,t}, \tilde{p}_{b,t}), 
\end{equation} 
where $\mathrm{sim}(\cdot,\cdot)$ denotes the distributional similarity function, which is instantiated using cosine similarity in this paper. During training, we minimize $\mathcal{L}_{\text{cv-dis}}$, thus explicitly encouraging differences in the output distributions across the two views.

It is important to emphasize that this cross-view difference optimization does not rely on the statistics of poisoned sample ratios at the batch level, but rather acts on the cross-view output alignment relation at the single-sample level. For normal samples, the outputs themselves are naturally diverse across views, so the gradient magnitude corresponding to this loss is limited, while for backdoor samples, the outputs collapse abnormally into almost identical fixed templates in both views, making $\mathrm{sim}(p_{b,t}, \tilde{p}_{b,t})$ significantly high, thus generating an abnormally amplified gradient signal under this loss. As a result, even under the \rev{low-to-moderate} poisoning rate setting of only $0\sim1$ poisoned samples in each batch, the regularization is still able to strengthen the constraints on backdoor behavior.

\subsection{Uncertainty-aware Output Regularization}
\label{subsec:regularization}
Although cross-view output difference optimization can effectively suppress the abnormal stability of the backdoor response, simply widening the output distribution difference under different views may induce the model to produce overconfident extreme predictions at some locations, which in turn leads to output distribution collapse or even degradation of normal generation capability. To avoid this side effect, we further introduce uncertainty-aware regularization constraints at the output layer~\cite{pereyra2017regularizing} to limit the degree of over-concentration of the model's predictions from the perspective of distributional entropy, in order to stabilize the training process and maintain the diversity of generation on normal samples.

For the output distribution $p_{b,t}$ of the $b$-th sample in the batch at the $t$-th token position, the information entropy is defined as:
\begin{equation} 
H_{b,t} = - \sum_{v \in \mathcal{V}} p_{b,t}^{(v)} \log p_{b,t}^{(v)}, 
\end{equation} 
where $p_{b,t}^{(v)}$ denotes the predicted probability of the $v$-th token in the vocabulary.

On this basis, we define the output entropy regularization term as:
\begin{equation} 
\mathcal{L}_{\text{ent}} 
= \frac{1}{|\Omega|} 
\sum_{(b,t)\in\Omega} 
\max(0, H_0 - H_{b,t}), 
\end{equation} 
where $H_0$ is a preset lower-bound threshold for entropy. This regularization term imposes penalties when the model produces over-concentrated low-entropy outputs at certain locations~\cite{guo2017calibration}, thus preventing the model from falling into an overconfident and probability-collapsing state.

The uncertainty-aware regularization term works synergistically with cross-view output discrepancy optimization to suppress the abnormal stability of the backdoor response on the one hand, and ensure that the model's generative distribution on normal samples maintains sufficient expressiveness and diversity on the other hand, so as to achieve a balance between security and normal generative capability.

\subsection{Implementation Details of Our Method}
\label{subsec:implementation}
Our approach is based on the standard supervised fine-tuning (SFT) process, where the original view $x_b$ and the \rev{patch-based perturbation} with a view $\tilde{x}_b$ are constructed synchronously for the same input samples in each training iteration and fed into the model for forward propagation at the same time to obtain the corresponding output distributions, respectively. The training objective consists of the task loss $\mathcal{L}_{\text{task}}$ together with three regularization terms $\mathcal{L}_{\text{patch}}$, $\mathcal{L}_{\text{cv-dis}}$, and $\mathcal{L}_{\text{ent}}$, and is jointly optimized by weighted summation:
\begin{equation} 
\mathcal{L}_{\text{def}} 
= \mathcal{L}_{\text{task}}
+ \lambda_1 \mathcal{L}_{\text{patch}}
+ \lambda_2 \mathcal{L}_{\text{cv-dis}}
+ \lambda_3 \mathcal{L}_{\text{ent}} .
\end{equation} 
where $\lambda_1, \lambda_2, \lambda_3$ are the weight coefficients of the different regularization terms.

\section{Experiments}\label{sec:experiments}

\subsection{Experimental Setup}

\textbf{Datasets.}
We evaluate the defense effectiveness of the proposed method on two widely used multimodal vision--language benchmark datasets, MS COCO and Flickr30k. Following standard evaluation protocols, we use the official test sets of these two datasets as clean test sets for evaluating the model's normal generation capability and visual comprehension performance under no-trigger conditions. In the backdoor training phase, we use LADD as the poisoning training dataset, and inject backdoor triggering patterns into a small number of samples under the 5\% poisoning ratio setting, which is used to construct the backdoor threat environment in the training phase.

\textbf{Threat Model and Victim Models.}
In order to verify the generality of the proposed defense approach under different multimodal large language model architectures, we select three representative models as victim models for evaluation, namely OpenFlamingo, Otter, and BLIP-2. These three types of models cover the general multimodal model based on cross-modal attention, the instruction-aligned multimodal model, and the bridged multimodal model, respectively, representing three typical paradigms of current mainstream multimodal large language models.

\textbf{Trigger and Attack Success Determination.}
In all backdoor attack experiments, we uniformly choose the word \texttt{banana} as the textual trigger target keyword. When the model includes the target trigger word \texttt{banana} in the generated result, the backdoor attack on the sample is judged successful. This definition does not rely on manual semantic judgment and provides clear and reproducible automated evaluation criteria.

\textbf{Evaluation Metrics.}
We adopt two text generation evaluation metrics, BLEU@4 (denoted as B@4) and CIDEr, to measure the model's normal image captioning and language generation capabilities on the clean test set, and the Attack Success Rate (ASR) to measure the effectiveness of the backdoor attack under the triggering condition. \rev{Attack success is defined as the generation of attacker-specified 
target behavior. In our experiments, this corresponds to the presence 
of the target keyword ``banana'' in the output, which provides an 
objective and reproducible criterion for automated evaluation. 
We acknowledge that this metric may not capture semantic variations 
of malicious outputs, which is discussed as a limitation.}

\textbf{Backdoor Attacks.}
In order to systematically evaluate the effectiveness and robustness of the proposed defense method under different triggering mechanisms, we select six representative backdoor attack methods for experimental comparison, specifically BadNets, Blended, LowFrequency, WaNet, InputAware, and DualKey. These attacks cover typical pixel-level local triggering, global blending triggering, frequency-domain triggering, geometric transformation triggering, input-adaptive triggering, and dual-key triggering backdoor paradigms, which are able to characterize the differences of multimodal backdoor attacks in triggering form, covertness, and stability from multiple dimensions, so as to comprehensively validate the versatility and robustness of the proposed defense method.

\begin{table*}[t]
\centering
\caption{Defense Performance of OpenFlamingo on COCO Dataset}
\label{tab1}
\begin{tabular}{lcccccc}
\toprule
\multirow{2}{*}{Attack} & \multicolumn{3}{c}{No Defense} & \multicolumn{3}{c}{Ours} \\
\cmidrule(lr){2-4} \cmidrule(lr){5-7}
 & B@4 & CIDEr & ASR & B@4 & CIDEr & ASR \\
\midrule
Clean        & 25.8 & 98.9 & 0.3  & 23.9 & 89.4 & 0.8  \\
BadNets      & 2.7  & 1.5  & 98.9 & 14.6 & 45.5 & 48.7 \\
Blended      & 2.0  & 1.9  & 98.5 & 13.7 & 41.3 & 58.6 \\
LowFrequency & 20.9 & 78.9 & 98.9 & 24.0 & 83.1 & 38.7 \\
WaNet        & 19.6 & 74.2 & 96.7 & 22.7 & 78.0 & 45.8 \\
InputAware   & 22.5 & 80.4 & 98.1 & 23.6 & 84.6 & 27.1 \\
DualKey      & 22.4 & 79.6 & 98.9 & 23.6 & 83.7 & 30.7 \\
\bottomrule
\end{tabular}
\end{table*}

\begin{table*}[t]
\centering
\caption{Defense Performance of Otter and BLIP-2 on COCO Dataset}
\label{tab2}
\begin{tabular}{lcccccc}
\toprule
\multirow{2}{*}{Attack} & \multicolumn{3}{c}{Otter} & \multicolumn{3}{c}{BLIP-2} \\
\cmidrule(lr){2-4} \cmidrule(lr){5-7}
 & B@4 & CIDEr & ASR & B@4 & CIDEr & ASR \\
\midrule
Clean        & 20.3 & 85.9 & 0.6  & 21.9 & 86.8 & 0.8  \\
BadNets      & 10.6 & 34.7 & 55.0 & 11.5 & 37.2 & 52.1 \\
Blended      & 9.6  & 31.5 & 60.4 & 10.9 & 34.0 & 58.1 \\
LowFrequency & 20.9 & 74.0 & 42.8 & 21.1 & 75.5 & 40.1 \\
WaNet        & 19.7 & 68.9 & 48.4 & 20.9 & 71.8 & 46.2 \\
InputAware   & 20.7 & 74.4 & 32.8 & 22.0 & 77.2 & 30.6 \\
DualKey      & 20.9 & 73.5 & 35.3 & 21.4 & 76.0 & 34.0 \\
\bottomrule
\end{tabular}
\end{table*}

\begin{table*}[t]
\centering
\caption{Defense Performance on Flickr30k Dataset}
\label{tab3}
\begin{tabular}{lcccccc}
\toprule
\multirow{2}{*}{Attack} & \multicolumn{3}{c}{No Defense} & \multicolumn{3}{c}{Ours} \\
\cmidrule(lr){2-4} \cmidrule(lr){5-7}
 & B@4 & CIDEr & ASR & B@4 & CIDEr & ASR \\
\midrule
Clean        & 22.1 & 86.6 & 0.3  & 20.9 & 78.5 & 0.5  \\
BadNets      & 1.8  & 1.1  & 98.2 & 12.5 & 38.6 & 53.0 \\
Blended      & 1.9  & 1.8  & 98.7 & 10.2 & 33.7 & 60.8 \\
LowFrequency & 19.7 & 68.9 & 97.1 & 22.0 & 74.4 & 42.6 \\
WaNet        & 18.8 & 64.4 & 95.4 & 20.7 & 69.4 & 48.4 \\
InputAware   & 20.9 & 72.4 & 98.1 & 22.4 & 78.4 & 32.3 \\
DualKey      & 19.1 & 71.2 & 98.6 & 21.1 & 77.0 & 35.8 \\
\bottomrule
\end{tabular}
\end{table*}

\begin{table}[t]
\centering
\caption{Ablation Study on BadNets Attack (COCO Dataset)}
\label{tab4}
\begin{tabular}{lccc}
\toprule
Method & B@4 & CIDEr & ASR \\
\midrule
No Defense                      & 2.0  & 1.5  & 98.1 \\
$L_{\text{patch}}$              & 8.2  & 22.9 & 65.6 \\
$L_{\text{patch}} + L_{\text{cv-dis}}$ & 12.3 & 35.5 & 53.7 \\
Ours                            & 14.6 & 45.5 & 48.7 \\
\bottomrule
\end{tabular}
\end{table}

\subsection{Defense Performance under Different Attacks}

Table~\ref{tab1} reports the comparison of the defense performance of different approaches on the COCO test set under six typical backdoor attack settings, where the evaluation metrics include the normal image description performance (B@4, CIDEr) and the backdoor attack success rate (ASR). From the Clean scenario, it can be seen that under clean inputs without triggers, our method introduces only a limited performance degradation on B@4 and CIDEr (B@4 decreases from 25.8 to 23.9, and CIDEr decreases from 98.9 to 89.4), which indicates that the proposed defense strategy has good stability in maintaining the normal generation capability.

In the backdoor attack scenario, the ASR of the undefended model under all six attack methods is close to the saturation level and reaches or exceeds 96\%, indicating that once the model is successfully implanted with a backdoor, it is almost inevitable to output the attack target under the triggering conditions, and it has extremely strong attack stability. In contrast, after the introduction of this paper's defense method, the ASRs of all attack methods are significantly suppressed, in which the ASRs of BadNets, Blended, LowFrequency, WaNet, InputAware, and DualKey are reduced from 98.9\%, 98.5\%, 98.9\%, 96.7\%, 98.1\%, and 98.9\% to 48.7\%, 58.6\%, 38.7\%, 45.8\%, 27.1\%, and 30.7\%, respectively. This result shows that the proposed method can stably weaken the trigger--target binding relationship under different triggering mechanisms, and has strong suppression ability against multiple types of backdoor attacks.

From the perspective of normal performance recovery, it can be seen that in most attack scenarios, our method not only significantly reduces the ASR, but also recovers the B@4 and CIDEr metrics to varying degrees. For example, under BadNets and Blended attacks, B@4 improves from 2.7 and 2.0 to 14.6 and 13.7, respectively, and CIDEr improves from 1.5 and 1.9, which are close to failure, to 45.5 and 41.3, which indicates that the normal generative capacity of the model is effectively restored after defense. Combining the above results, it can be concluded that under the six different backdoor attack paradigms, this paper's method is able to maintain the normal image generation capability of the model while significantly suppressing the success rate of the attacks, which verifies the effectiveness and stability of the proposed defense framework in realistic low-poisoning and multi-attack scenarios.

\subsection{Defense Performance across Different Models}

In order to evaluate the generalization ability of the proposed defense method under different multimodal large language model architectures, we further conducted experimental validation on two representative models, Otter and BLIP-2, and the results are shown in Table~\ref{tab2}. It can be seen that in the Clean scenario without triggers, both models maintain a relatively stable normal generation performance after the introduction of defenses, with Otter reaching 20.3 and 85.9 on B@4 and CIDEr, respectively, and BLIP-2 reaching 21.9 and 86.8, with an ASR close to 0. This indicates that the proposed method does not significantly disrupt the normal image generation capability under different model architectures.

In the backdoor attack scenario, the ASRs of Otter and BLIP-2 are significantly suppressed under all six attack methods. Taking Otter as an example, the ASRs under BadNets, Blended, LowFrequency, WaNet, InputAware, and DualKey attacks are 55.0\%, 60.4\%, 42.8\%, 48.4\%, 32.8\%, and 35.3\%, respectively; on BLIP-2, the ASRs are further reduced to 52.1\%, 58.1\%, 40.1\%, 46.2\%, 30.6\%, and 34.0\%. Compared with the attack success rate in the undefended state, which is generally close to saturation, the ASRs in the defended state are all significantly reduced, indicating that the method can effectively weaken the trigger--target binding relationship under different model architectures.

In terms of normal performance recovery, the B@4 and CIDEr metrics remain close to the Clean scenario in most attack settings. For example, under LowFrequency and InputAware attacks, the B@4 of Otter and BLIP-2 remains around 20, and the CIDEr is maintained in the range of 70--77, which indicates that the defense strategy is able to suppress the backdoor behaviors while simultaneously retaining the normal generation capability of the model.

Taking the above results together, it can be concluded that the proposed defense method is not only effective on a single model, but also shows stable defense performance and good normal performance preservation under different multimodal model architectures, such as Otter and BLIP-2, which verifies the good generalizability of the method at the model level.

\subsection{Defense Performance across Different Datasets}

To further validate the generalization ability of the proposed defense method under different data distributions, we evaluate the model on the Flickr30k dataset, and the experimental results are shown in Table~\ref{tab3}. It can be seen that in the Clean scenario, the normal generation performance of the model only shows a limited degradation after the introduction of the defense, with B@4 decreasing from 22.1 to 20.9, CIDEr decreasing from 86.6 to 78.5, and ASR remaining close to 0. This indicates that the proposed method can also maintain the normal image generation capability of the model under cross-dataset scenarios.

In the backdoor attack scenario, the ASRs of the undefended model under the six attack methods are still generally close to saturation, and all of them reach more than 95\%, which indicates that the backdoor behavior under the Flickr30k data distribution is also extremely stable and harmful. In contrast, after applying the proposed defense method, the ASRs under BadNets, Blended, LowFrequency, WaNet, InputAware, and DualKey attacks are reduced to 53.0\%, 60.8\%, 42.6\%, 48.4\%, 32.3\%, and 35.8\%, respectively, showing a consistent and stable downward trend overall.

From the perspective of normal performance recovery, in most attack scenarios, the B@4 and CIDEr metrics are significantly improved after defense. For example, under BadNets and Blended attacks, B@4 improves from 1.8 and 1.9, which are close to failure, to 12.5 and 10.2, respectively, and CIDEr improves from 1.1 and 1.8 to 38.6 and 33.7, which indicates that the model's normal generation capability is effectively restored after defense.

Combining the results in Table~\ref{tab3}, it can be concluded that the proposed defense method is not only effective on the COCO dataset, but also can significantly reduce the success rate of backdoor attacks and maintain a more stable normal generation performance on Flickr30k, a test set with different data distributions, which verifies the generalization ability and robustness of the proposed method in cross-dataset scenarios.

\subsection{Ablation Study}

To further analyze the role of each regularization term in the defense framework, we conduct ablation experiments under the BadNets attack setting on the COCO dataset, and the results are shown in Table~\ref{tab4}. The experiments sequentially compare No Defense, the introduction of only the feature consistency regularization $\mathcal{L}_{\text{patch}}$, the simultaneous introduction of $\mathcal{L}_{\text{patch}}$ with the cross-view output difference regularization $\mathcal{L}_{\text{cv-dis}}$, and the performance variation of the complete method (Ours).

In terms of ASR, the undefended model has an ASR as high as 98.1\%, indicating that the backdoor attack is almost stably effective under the triggering condition. When only $\mathcal{L}_{\text{patch}}$ is introduced, the ASR drops significantly to 65.6\%, indicating that the block-level feature consistency constraints are able to weaken the model's overfitting to localized triggering patterns to a certain extent, but it is still difficult to completely disrupt the trigger--target binding relationship. After further introducing the cross-view output difference regularization $\mathcal{L}_{\text{cv-dis}}$, the ASR further decreases to 53.7\%, indicating that constraining cross-view anomalous invariance at the output distribution level is a key factor in suppressing backdoor triggering behavior. Ultimately, the complete method further suppresses the ASR to 48.7\% after introducing the three losses simultaneously, achieving the optimal defense effect.

From the perspective of normal generation performance, B@4 and CIDEr show a steady increase with the gradual introduction of the defense modules. Compared to the almost failed generation performance without defense (2.0 for B@4 and 1.5 for CIDEr), the scores can be significantly recovered to 8.2 and 22.9 with the introduction of $\mathcal{L}_{\text{patch}}$ alone, and are further improved to 12.3 and 35.5 with the stacking of $\mathcal{L}_{\text{cv-dis}}$; the complete method finally reaches 14.6 and 45.5, which is close to the performance level of the normal model in the Clean scenario. This suggests that the proposed three regularization terms suppress backdoor behavior and at the same time promote the restoration of the model's normal generation capability.

Combining the above results, it can be concluded that $\mathcal{L}_{\text{patch}}$ is mainly responsible for weakening the model's feature dependence on local trigger patterns, $\mathcal{L}_{\text{cv-dis}}$ is the core constraint that directly breaks the trigger--target binding relationship and reduces the ASR, and $\mathcal{L}_{\text{ent}}$ further stabilizes the output distribution and improves the overall generation performance in the complete framework. All three synergistically form an essential part of the defense framework in this paper.

\subsection{Hyper-parameter Analysis}
\begin{figure}[htbp]
\centering
\includegraphics[width=1.0\columnwidth]{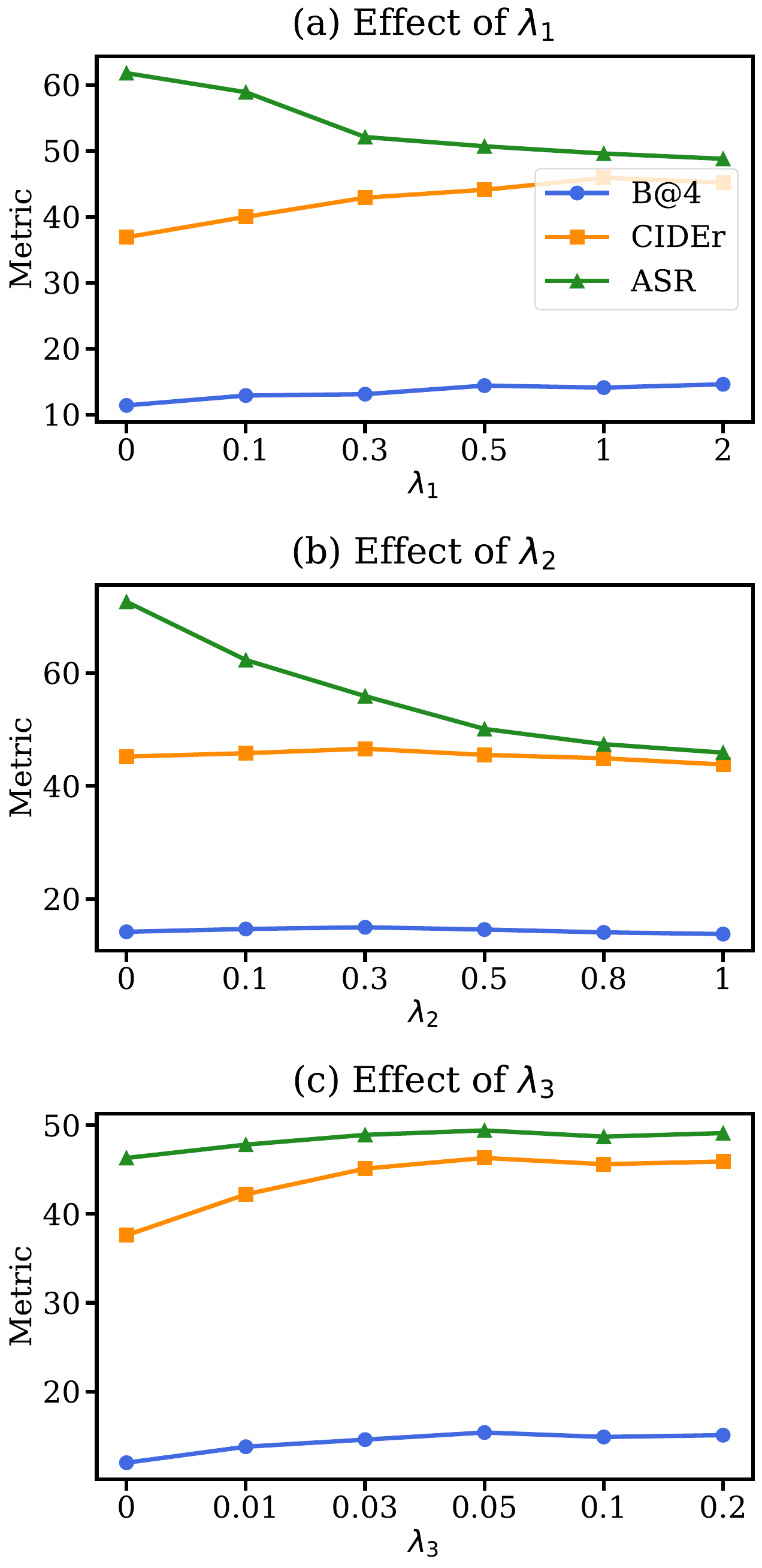}
\caption{Impact of different regularization weights $\lambda_1$, $\lambda_2$, and $\lambda_3$ on the trade-off between normal generation performance (B@4, CIDEr) and ASR.}
\label{fig:abl}
\end{figure}

In order to analyze the effects of different regularization weights on the defense performance and normal generation capability, we conduct hyper-parameter sensitivity experiments on $\lambda_1$ (feature consistency regularization), $\lambda_2$ (cross-view output discrepancy regularization), and $\lambda_3$ (uncertainty-aware entropy regularization), respectively, and the results are shown in Figure~\ref{fig:abl}.

Firstly, it can be observed from Figure~\ref{fig:abl} (a) that with the gradual increase of $\lambda_1$, the ASR of the model shows an overall decreasing trend, while B@4 and CIDEr gradually increase and stabilize. This suggests that appropriately enhancing the block-level feature consistency constraints can help weaken the model's overfitting to local triggering patterns without significantly impairing the normal generation capability. When $\lambda_1$ is too large, the performance improvement tends to saturate, indicating that this regularization term has a smooth impact on the overall performance within a reasonable range.

Secondly, Figure~\ref{fig:abl} (b) shows that $\lambda_2$ has the most direct and significant inhibitory effect on ASR. As $\lambda_2$ increases, ASR almost monotonically decreases, which verifies that the cross-view output difference regularization is the core constraint to break the trigger--target binding relationship and suppress the backdoor response. Meanwhile, B@4 and CIDEr remain relatively stable over a wide range, only slightly decreasing at larger weights, indicating that this regularization term can better maintain normal generation performance while enhancing security.

Finally, from Figure~\ref{fig:abl} (c), it can be observed that $\lambda_3$ can improve B@4 and CIDEr and maintain a low ASR within a small range of values, which indicates that the uncertainty-aware entropy regularization can help stabilize the output distribution and alleviate the over-confidence problem caused by the optimization of cross-view differences. When $\lambda_3$ is too large, the performance gain tends to flatten out or even fluctuate slightly, indicating that the entropy regularization is more suitable to be used as a stabilizing term rather than a dominant constraint.

Combining the above results, it can be concluded that the proposed defense method can maintain a stable defense effect and normal performance trade-off over a wide range of hyperparameter values, which demonstrates that the method is insensitive to hyperparameter settings and has good training stability and practical scalability.

\subsection{Comparison with Existing Defenses}
\begin{table}[t]
\centering
\caption{Defense performance against the BadNets attack with baselines.}
\label{tab_badnets}
\begin{tabular}{lccc}
\toprule
Defense Method & B@4 & CIDEr & ASR \\
\midrule
No Defense & 1.8 & 1.1 & 98.2 \\
Fine-Pruning & 9.6 & 29.4 & 71.3 \\
STRIP Detection & 3.2 & 6.8 & 88.5 \\
Entropy Filtering & 7.4 & 21.7 & 79.6 \\
\midrule
\textbf{Ours} & \textbf{12.5} & \textbf{38.6} & \textbf{53.0} \\
\bottomrule
\end{tabular}
\end{table}
\rev{The results of the BadNets attack experiments performed on the Flickr30k dataset are shown in Table~\ref{tab_badnets}. It can be seen that without any defense measures, the model is almost completely controlled by the backdoor, and the ASR is as high as 98.2\%, while the generation quality is severely degraded, with BLEU-4 and CIDEr being only 1.8 and 1.1, respectively, which indicates that the model can no longer properly complete the description task under the triggering conditions.}

\rev{In contrast, existing defense methods in the inference phase can only partially mitigate the attack. Fine-Pruning~\cite{liu2018fine} reduces the ASR to 71.3\%, but still fails to effectively prevent backdoor behavior, and the generation performance is still low. STRIP~\cite{gao2019strip} detection methods have limited protection against this visual trigger, and the ASR is still as high as 88.5\%, which indicates that consistency detection based on input perturbation is difficult to identify this type of multimodal trigger. The entropy filtering~\cite{chen2018detecting} method performs in between, with an ASR of 79.6\%, but at the same time significantly affects the quality of the model output.}

\rev{In addition, this paper’s method significantly reduces the attack success rate, lowering the ASR to 53.0\%, while maintaining higher generation quality (12.5 for BLEU-4 and 38.6 for CIDEr). This suggests that, unlike inference-phase defense that relies on post-hoc detection or pruning, we are able to more effectively improve the robustness of the model and strike a better balance between security and task performance by suppressing the formation of backdoor mechanisms during the training phase.}

\subsection{Computational Overhead Analysis}
\rev{Our method introduces an additional forward pass for the perturbed view 
during training, resulting in approximately a twofold increase in 
training-time computation compared to standard SFT.
In practice, GPU memory usage increases moderately due to the storage 
of intermediate activations for both views. Importantly, the method 
does not introduce any additional overhead during inference, as the 
perturbation mechanism is only applied in the training phase.
Therefore, the framework is suitable for scenarios where training-time 
cost is acceptable but inference efficiency is critical.
}

\subsection{Discussion}
\rev{
Overall, the proposed method demonstrates consistent defense performance 
across diverse attack types, datasets, and model architectures. 
The results suggest that cross-view discrepancy regularization 
effectively disrupts trigger–target bindings while preserving 
normal generation behavior.}

\rev{
However, performance gains vary across attacks. Methods relying on 
global or adaptive triggers (e.g., Blended or InputAware) remain 
more challenging to suppress, indicating potential limitations 
of localized perturbation-based defenses.}

\section{Conclusion and Limitations}\label{sec5}
In this paper, we propose a multimodal backdoor defense framework based on block-level augmentation and cross-view regularity, which characterizes and suppresses the anomalous invariance properties of backdoor responses under non-semantic perturbations from the geometric structure of the cross-view output distribution. Through the synergistic constraints of block-level view generation, cross-view output discrepancy optimization, and uncertainty-aware output regularization, the proposed method is able to stably reduce the attack success rate under very low poisoning ratios or even single-sample-level backdoor scenarios, while maintaining the normal generation capability of the model, and its effectiveness and generalizability are verified under multi-model, multi-task, and multi-attack settings. This work provides a mechanistic pathway for multimodal backdoor defense that is decoupled from the attack scale. Meanwhile, this method still has some limitations: its core perturbation mainly operates on localized visual regions, and further evaluation is needed for more complex forms of cross-modal synergistic triggering; the regularization weights under different models and tasks still need to be adjusted; and the method mainly relies on the fine-tuning phase to take effect, with limited direct applicability to already deployed models. In the future, we will explore the extension to more complex triggering modes, adaptive weight scheduling, and lightweight protection mechanisms in the inference phase.

\bmhead{Acknowledgements}
The authors would like to thank all collaborators and institutions that provided open-source models, datasets, and toolkits used in this research.

\bmhead{Author contributions}
All authors contributed to the conceptualization, methodology design, implementation, experimental evaluation, and manuscript preparation of this work.

\bmhead{Data availability}
All datasets used in this paper are publicly available. For reproducibility, the processed data and experimental protocols will be released upon acceptance of this paper.

\bmhead{Code availability}
The code will not be released during the review stage due to ongoing extension of this work.

\section*{Declarations}
\textbf{Conflict of interest} The authors declare no competing interests.

\nocite{*}
\bibliography{sn-bibliography}

\end{document}